\pdfoutput=1
\documentclass{article}
\usepackage{spconf,amsmath,amssymb,graphicx}
\usepackage{algorithm}
\usepackage[noEnd=true,indLines=true]{algpseudocodex}
\algrenewcommand{\algorithmicrequire}{\textbf{Input:}}
\algrenewcommand{\algorithmicensure}{\textbf{Output:}}
\usepackage{enumitem}
\setlist{nosep,leftmargin=14pt}
\usepackage{booktabs,placeins}
\usepackage[hidelinks]{hyperref}
\newcommand{\ImgEst}{\operatorname{ImgEst}}
\newcommand{\MotEst}{\operatorname{MotEst}}
\renewcommand{\dblfloatpagefraction}{0.8}
\makeatletter
\renewcommand{\section}{\@startsection{section}{1}{\z@}%
  {-2.2ex plus -.5ex minus -.2ex}{1.1ex plus .2ex}%
  {\normalfont\normalsize\bfseries}}
\renewcommand{\subsection}{\@startsection{subsection}{2}{\z@}%
  {-1.7ex plus -.4ex minus -.2ex}{.7ex plus .2ex}%
  {\normalfont\normalsize\bfseries}}
\long\def\@makecaption#1#2{%
  \vskip 4pt
  \setbox\@tempboxa\hbox{#1. #2}%
  \ifdim\wd\@tempboxa>\hsize #1. #2\par
  \else\hbox to\hsize{\hfil\box\@tempboxa\hfil}\fi}
\let\isbi@thebibliography\thebibliography
\renewcommand{\thebibliography}[1]{%
  \isbi@thebibliography{#1}%
  \setlength{\itemsep}{1pt}%
  \setlength{\parsep}{0pt}}
\makeatother
 \newif\ifmethodsonly
\methodsonlyfalse
\title{Training-Free Motion-Corrected 3D Brain MRI Reconstruction via Event-Guided References}
\name{Wei Jiang$^{1}$, Hongli Chen$^{1}$, Tianyi Ding$^{2}$, Feng Liu$^{1}$, Nan Ye$^{1}$, Hongfu Sun$^{3}$}
\address{$^{1}$The University of Queensland, Brisbane, Australia\\
$^{2}$RMIT University, Melbourne, Australia\\
$^{3}$The University of Newcastle, Newcastle, Australia}
\begin{document}
\pagestyle{plain}
\maketitle
\suppressfloats[t]
\begin{abstract}
Head motion causes artefacts in 3D brain MRI and can bias structural measurements. Model-based correction alternates between image and motion estimation, but artefacts in the reconstructed reference can reinforce motion errors. Learned priors can reduce these artefacts but require pretraining and may add test-time computation. We propose EventMC (event-guided motion correction), a training-free method for interleaved Cartesian acquisitions. It iteratively detects motion events and estimates motion using reference images reconstructed from stable intervals. It then refines the image and motion from low to high spatial frequencies. In a fair comparison in simulation (same reconstruction), EventMC's motion estimates improve mean PSNR by up to 4.0~dB over those of the learning-based MotionTTT. A single run of EventMC is up to 7.5 times faster than MotionTTT with less than half the GPU memory. On in vivo motion-corrupted scans, EventMC achieves similar mean image quality to MotionTTT without any training. Code is available at \url{https://github.com/bayesbreeze/EventMC}.

\end{abstract}


\begin{figure}[t]
\centering
\includegraphics[width=\columnwidth]{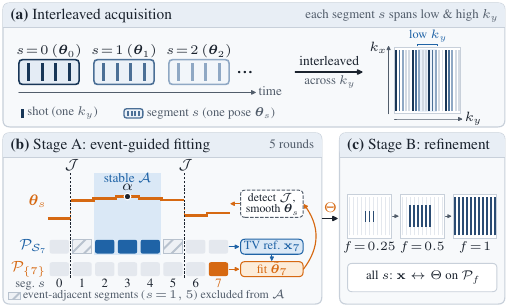}
\caption{EventMC overview (schematic). (a) Interleaved acquisition: consecutive shots (bars, one $k_y$ each) form segments (boxes, one pose $\boldsymbol{\theta}_s$ each) spread over $k_y$. (b) Stage A, fitting segment 7 (orange, $\mathcal{P}_{\lbrace 7\rbrace}$) to a TV reference $\mathbf{x}_7$ from the stable interval $\mathcal{A}$ around the anchor $\alpha$ (blue, $\mathcal{P}_{\mathcal{S}_7}$; hatched: excluded event-adjacent segments); the poses (orange line) jump at events $\mathcal{J}$. (c) Stage B, coarse-to-fine refinement: all segments, with the central band $\mathcal{P}_f$ widening with $f$.}
\label{fig:method-overview}
\end{figure}
 
\section{Introduction}

Head motion causes blurring and ghosting artefacts in 3D brain MRI and can bias structural measurements. These artefacts can lead to underestimation of cortical thickness and grey-matter volume~\cite{reuter2015motion}, potentially confounding clinical comparisons~\cite{yao2017motion}. Motion correction is therefore important for image quality and structural analysis.

Model-based correction alternates between estimating the image and the motion~\cite{cordero2016,disorder,smeir}. Images are reconstructed from all segments with handcrafted priors such as total variation (TV)~\cite{rof}. When motion estimates are inaccurate, residual artefacts can bias subsequent pose fitting. Including the target segment in its own reference can reinforce these errors.

Learning-based methods use pretrained networks to suppress artefacts or guide reconstruction. Some, such as Stacked U-Net~\cite{almasni2022}, learn from paired motion-corrupted and motion-free images. Others learn from motion-free images alone. MotionTTT~\cite{motionttt} uses a U-Net reconstruction network to guide motion estimation at test time, while diffusion-based methods~\cite{motiondps,levac2024} build on diffusion reconstruction~\cite{jalal2021,chung2022,jiang2024fast}. All of these require pretraining, and MotionTTT's test-time training or iterative diffusion sampling adds further computation.

We propose EventMC, a training-free method for interleaved 3D Cartesian acquisitions. It uses stable intervals between abrupt motion events~\cite{zaitsev2015,qin2009} to build TV references from segments that each cover low and high spatial frequencies. Our method has two stages. Stage A iteratively detects events from pose jumps and estimates each segment's motion against a reference that excludes its own data. Stage B then refines the image and motion with all segments, from low to high spatial frequencies. We evaluate EventMC on simulated CC359 data~\cite{cc359} and 26 in vivo motion-corrupted PMoC3D scans~\cite{pmoc3d}. Our contributions are:
\begin{itemize}
\item \textbf{Event-guided references:} detected events select stable segments to form each TV reference, excluding the segment being fitted to reduce error reinforcement.
\item \textbf{Two-stage optimisation:} event-guided initialisation followed by low-to-high frequency refinement helps avoid the poor solutions of plain alternation.
\item \textbf{Training-free performance:} up to 4.0~dB higher mean PSNR than MotionTTT in simulation with the same reconstruction; up to 7.5 times faster with less than half the GPU memory in a single run; and similar mean image quality on real motion.
\end{itemize}
\ifmethodsonly
\begin{figure}[t]
\centering
\includegraphics[width=\columnwidth]{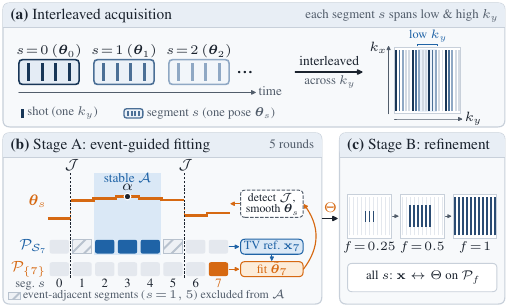}
\caption{EventMC overview (schematic). (a) Interleaved acquisition: consecutive shots (bars, one $k_y$ each) form segments (boxes, one pose $\boldsymbol{\theta}_s$ each) spread over $k_y$. (b) Stage A, fitting segment 7 (orange, $\mathcal{P}_{\lbrace 7\rbrace}$) to a TV reference $\mathbf{x}_7$ from the stable interval $\mathcal{A}$ around the anchor $\alpha$ (blue, $\mathcal{P}_{\mathcal{S}_7}$; hatched: excluded event-adjacent segments); the poses (orange line) jump at events $\mathcal{J}$. (c) Stage B, coarse-to-fine refinement: all segments, with the central band $\mathcal{P}_f$ widening with $f$.}
\label{fig:method-overview}
\end{figure}
 \fi

\section{Problem formulation}

We consider interleaved 3D Cartesian MRI whose shots are grouped into $N_s$ time-contiguous segments, $s=0,\ldots,N_s-1$, each sampling both low and high spatial frequencies~\cite{disorder} (Fig.~\ref{fig:method-overview}a). The multi-coil k-space data $\mathbf{y}$ are modelled as

\begin{equation}
\mathbf{y}=A(\Theta)\mathbf{x}^{*}+\mathbf{z},
\label{eq:forward}
\end{equation}

where $\mathbf{x}^{*}$ is the motion-free image, $\mathbf{z}$ is noise, and the motion trajectory $\Theta$ stacks one rigid pose per segment, $\boldsymbol{\theta}_s=(\mathbf{r}_s,\mathbf{t}_s)\in\mathbb{R}^6$ (rotation $\mathbf{r}_s$ in degrees, translation $\mathbf{t}_s$ in mm). For coil $c$, segment $s$ is encoded as $[A_s(\boldsymbol{\theta}_s)\mathbf{x}^{*}]_c=\mathcal{M}_s\mathcal{F}S_c\mathcal{T}(\boldsymbol{\theta}_s)\mathbf{x}^{*}$, with rigid transform $\mathcal{T}$ about the FOV centre, scanner-fixed coil sensitivity $S_c$, 3D Fourier transform $\mathcal{F}$ and the sampling mask $\mathcal{M}_s$ of segment $s$. We estimate $\mathbf{x}$ and $\Theta$ by minimising

\begin{equation}
\mathcal{L}(\mathbf{x},\Theta;\mathcal{P},\mathcal{R})
=\tfrac{1}{2}\Vert\mathcal{P}(A(\Theta)\mathbf{x}-\mathbf{y})\Vert_2^2
+\lambda\mathcal{R}(\mathbf{x}),
\label{eq:objective}
\end{equation}

where the \emph{k-space selector} $\mathcal{P}$ keeps part of the acquired samples ($\mathcal{P}_{\mathcal{S}}$: segments in $\mathcal{S}$; $\mathcal{P}_f$: the central fraction $f$ of each k-space axis; $I$: all samples), and $\mathcal{R}$ is an image regulariser weighted by $\lambda$; image updates use smoothed isotropic total variation~\cite{rof}, $\mathcal{R}=\operatorname{TV}$.

\section{Method}

\ifmethodsonly\else
\begin{figure*}[!t]
\centering
\includegraphics[width=\textwidth]{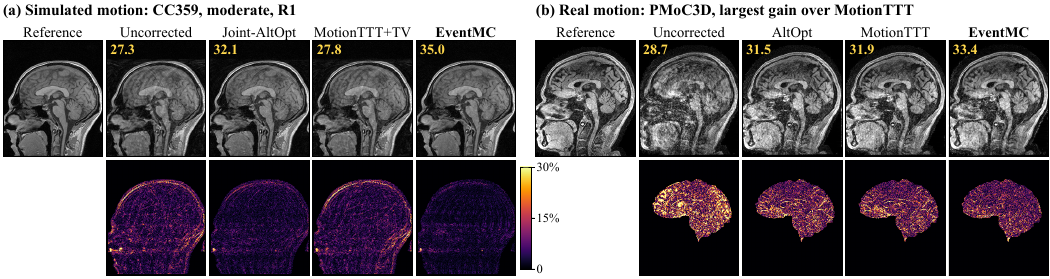}
\caption{Example reconstructions under (a) simulated and (b) real motion; baselines in (b) are the released reconstructions. Top: central sagittal slices with volumetric PSNR (dB); bottom: absolute errors relative to the reference display range, within the head mask and valid FOV (a) or brain mask (b).}
\label{fig:cases}
\end{figure*}
 \fi

Rather than fitting motion against an image reconstructed from all segments, EventMC estimates each segment's motion against a reference image with fewer artefacts, reconstructed from a stable interval between detected motion events (Fig.~\ref{fig:method-overview}b). Specifically, a pre-run selects an anchor $\alpha$; Stage A alternates between updating these references and the motion, and Stage B refines both with all segments, from low to high spatial frequencies (Algorithm~\ref{alg:single-anchor}). All steps share two updates of Eq.~\eqref{eq:objective}: $\ImgEst(\mathbf{x},\allowbreak \Theta;\allowbreak \mathcal{P}^{\mathrm{img}})$ and $\MotEst(\mathbf{x},\allowbreak \Theta;\allowbreak \mathcal{P}^{\mathrm{mot}})$ approximately solve $\min_{\mathbf{x}}\mathcal{L}(\mathbf{x},\allowbreak \Theta;\allowbreak \mathcal{P}^{\mathrm{img}},\allowbreak \operatorname{TV})$ and $\min_{\Theta}\mathcal{L}(\mathbf{x},\allowbreak \Theta;\allowbreak \mathcal{P}^{\mathrm{mot}},\allowbreak 0)$, respectively, starting from the given $\mathbf{x}$ or $\Theta$ with the other kept fixed; $0$ is the zero regulariser. Both use L-BFGS~\cite{lbfgs} with automatic differentiation; rotations are applied by Fourier shears~\cite{unser1995} and translations by phase ramps.

An \emph{event} $j\in\mathcal{J}$ is a detected jump between segments $j$ and $j+1$; events divide the scan into \emph{intervals} of consecutive segments. The \emph{anchor} $\alpha$ is the segment whose pose is fixed at $\boldsymbol{\theta}_{\alpha}=\mathbf{0}$. The steps differ mainly in their k-space selectors $\mathcal{P}^{\mathrm{img}}$ and $\mathcal{P}^{\mathrm{mot}}$.

\begin{algorithm}[t]
\caption{EventMC-S: single-anchor estimation}
\label{alg:single-anchor}
\small
\begin{algorithmic}[1]
\Require $\mathbf{y}$, sampling information, coil sensitivities
\Ensure $\hat{\mathbf{x}},\hat{\Theta}$
\State \textbf{Initialise:} form uncorrected $\mathbf{x}_0$ from $\mathbf{y}$
\State $\alpha\gets$ pre-run, $\Theta\gets\mathbf{0}$, $\mathcal{J}\gets\emptyset$, $\mathbf{x},\mathbf{x}_s\gets\mathbf{x}_0$
\For{5 rounds} \Comment{Stage A}
    \For{$s\ne\alpha$, outwards from $\alpha$}
        \State Set $\mathcal{P}^{\mathrm{img}}_s,\mathcal{P}^{\mathrm{mot}}_s$ \Comment{Eq.~\eqref{eq:selection}}
        \State $\mathbf{x}_s\gets\ImgEst(\mathbf{x}_s,\Theta;\mathcal{P}^{\mathrm{img}}_s)$
        \State $\tilde{\boldsymbol{\theta}}_s\gets\MotEst(\mathbf{x}_s,\boldsymbol{\theta}_s;\mathcal{P}^{\mathrm{mot}}_s)$ \Comment{$\boldsymbol{\theta}_s\in\Theta$}
    \EndFor
    \State Detect $\mathcal{J}$ from $\tilde{\Theta}$; smooth $\tilde{\Theta}\rightarrow\Theta$
\EndFor
\For{$f=0.25,0.5,1,1$} \Comment{Stage B, from $\mathbf{x}=\mathbf{x}_0$}
    \State Set $\mathcal{P}_f$ \Comment{All segments}
    \State $\mathbf{x}\gets\ImgEst(\mathbf{x},\Theta;\mathcal{P}_f)$
    \State $\Theta\gets\MotEst(\mathbf{x},\Theta;\mathcal{P}_f)$
\EndFor
\State $\mathbf{x}\gets\ImgEst(\mathbf{x}_0,\Theta;I)$
\State \Return $(\hat{\mathbf{x}},\hat{\Theta})=(\mathbf{x},\Theta)$
\end{algorithmic}
\end{algorithm}

\subsection{Stage A: fitting to event-guided references}
\label{sec:stage-a}

\textbf{Pre-run.} EventMC first selects $\alpha$ within a long stable part of the scan (Fig.~\ref{fig:method-overview}b). It runs one Stage A round from zero motion with the temporary anchor $\alpha_0=\lfloor N_s/2\rfloor$ and detects events. After excluding segments adjacent to an event or with likely motion within them (high sub-window prediction residuals; CC359 only), it takes as $\alpha$ the middle remaining segment of the interval with the most remaining segments.

\textbf{Reference segments.} The segments used to build the reference image $\mathbf{x}_s$ must be chosen carefully, since fewer artefacts in $\mathbf{x}_s$ improve the motion estimate of target segment $s$. To minimise head movement in the chosen segments, we use the stable interval $\mathcal{A}$ (Fig.~\ref{fig:method-overview}b): the anchor's interval without the segments next to an event, which may contain the actual movement. The reference for target $s$ is built from
\begin{equation}
\mathcal{S}_s=\begin{cases}\mathcal{A}, & s\notin\mathcal{A},\\ \lbrace s'\in\mathcal{A}:s'\bmod4\ne s\bmod4\rbrace, & s\in\mathcal{A},\end{cases}
\label{eq:selection}
\end{equation}
so it never contains the target's own data, and grouping by $s\bmod4$ lets targets share references. Thus $\mathcal{P}^{\mathrm{img}}_s=\mathcal{P}_{\mathcal{S}_s}$ and $\mathcal{P}^{\mathrm{mot}}_s=\mathcal{P}_{\lbrace s\rbrace}$, both at all frequencies. With no events yet in Round 1, $\mathcal{A}$ contains all segments; each reference starts from $\mathbf{x}_0$ or from the previous reference with the same omitted group.

\textbf{Pose fitting and events.} $\MotEst$ fits the pose of each target segment, from the nearest to $\alpha$ to the farthest~\cite{namer}, so each fit can start from the already fitted pose of its neighbour closer to $\alpha$. Since the fit is non-convex, each fit is run from three starting poses: the previous-round pose, the neighbour's pose, and the neighbour's pose plus the previous-round difference between the two segments; the result with the lowest residual is kept. Events are then detected as jumps between adjacent poses: $j$ is an event if the root-mean-square displacement $m_j$~\cite{jenkinson} between segments $j$ and $j+1$ exceeds $\max(3\sigma_j,0.6\,\mathrm{mm})$, with $\sigma_j$ a robust local noise level. Finally, the raw fits $\tilde{\Theta}$ are smoothed by balancing fidelity to them against a Huber penalty on their second temporal differences~\cite{motiondps}, except across events so that jumps are preserved. The next round uses $\mathcal{J}$ to select references and $\Theta$ for encoding and initialisation.

\subsection{Stage B: coarse-to-fine refinement}
\label{sec:stage-b}

Stage B refines the motion and one shared image $\mathbf{x}$ using all segments, from coarse to fine. Starting from the Stage A motion and the uncorrected image, it alternates $\ImgEst$ and $\MotEst$ on the central fraction $f$ of each k-space axis ($\mathcal{P}^{\mathrm{img}}=\mathcal{P}^{\mathrm{mot}}=\mathcal{P}_f$, Fig.~\ref{fig:method-overview}c), with $f$ increasing as $0.25,0.5,1,1$, since low frequencies are less sensitive to small pose errors. Finally, $\mathbf{x}$ is reconstructed from all samples ($\mathcal{P}=I$) with $\Theta$ fixed, starting again from the uncorrected image.

\subsection{Multi-anchor extension}
\label{sec:multi-anchor}

Since a single run of Algorithm~\ref{alg:single-anchor} (EventMC-S) can fail with a poorly placed anchor, EventMC runs up to three candidates and keeps the most data-consistent one. The candidates differ only in Stage A, which runs (i) with $\alpha$ at full resolution, as in EventMC-S, (ii) with $\alpha$ at half resolution, or (iii) with a second anchor $\alpha_2$ at half resolution. Half resolution uses only the central half of each k-space axis, with re-estimated coil sensitivities. The pre-run picks $\alpha_2$ by the same rule from the interval with the second-most remaining segments, if such an interval exists. All candidates then run Stage B and the final reconstruction at full resolution, with poses carried over unchanged in physical units. EventMC keeps the candidate with the smallest full-data residual $\Vert A(\Theta)\mathbf{x}-\mathbf{y}\Vert_2^2$; this is a heuristic, as a smaller residual does not guarantee a smaller image error.
\section{Experiments}

\begin{figure}[t]
\centering
\includegraphics[width=\columnwidth]{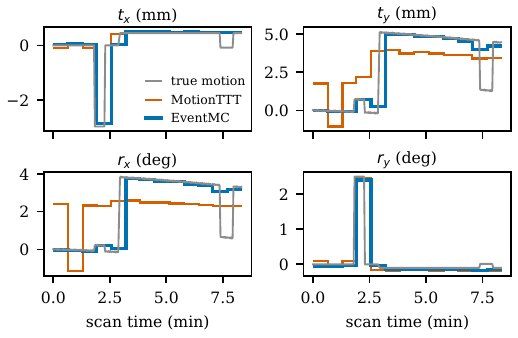}
\caption{True and estimated motion for one CC359 subject (moderate, R1; translations $t_x,t_y$ and rotations $r_x,r_y$). Each method estimates one pose per segment (piecewise constant).}
\label{fig:traj}
\end{figure}

\subsection{Experimental setup}

\textbf{Data.} We evaluate on simulated and real head motion, both with 13 interleaved segments. For simulation, we add rigid motion to the 12-coil 3D k-space of the nine CC359 subjects selected according to MotionTTT's test split~\cite{motionttt,cc359}; each scan has 218 shots of 2.3~s, each acquiring one $k_y$ index ($k_y\equiv5s\pmod{13}$ in segment $s$). Following reported head-motion characteristics~\cite{zaitsev2015,qin2009}, the motion combines slow drift with one to three abrupt events (\emph{moderate}) or one sustained mid-scan event (\emph{M}), and drift alone (\emph{subthreshold}) serves as a control. Each motion is applied at full sampling (R1) and at two- and four-fold Poisson-disc undersampling (R2, R4; MotionTTT's mask at R4), giving 54 moderate and M cases. For real motion, we use 26 of the 27 motion scans of PMoC3D~\cite{pmoc3d} (eight subjects, $R\approx4.9$, 52 shots, four per segment), each with the subject's motion-free scan as reference; one scan is excluded for a coil-count mismatch.

\textbf{Baselines and evaluation.} For simulation, we compare with MotionTTT (official code and weights)\footnote{MotionTTT's Stacked U-Net~\cite{almasni2022}, trained on paired data from a different acquisition, stays close to Uncorrected here.}, MotionTTT+TV (its motion with our final TV reconstruction, Alg.~\ref{alg:single-anchor}, line~13, so that both share the final step; unlike MotionTTT's forward model, ours keeps coil sensitivities scanner-fixed) and Joint-AltOpt (plain alternation of Eq.~\eqref{eq:objective} on all data~\cite{disorder}). For real motion, we compare with the released MotionTTT, AltOpt~\cite{cordero2016}, Stacked U-Net~\cite{almasni2022} and uncorrected L1 (Uncorrected) reconstructions. PSNR and SSIM are computed after rigid alignment within the head mask and valid FOV for simulation, and with the MotionDPS protocol~\cite{motiondps} for real motion (registration, FastSurfer brain mask~\cite{fastsurfer} and intensity normalisation). Two-sided Wilcoxon tests use per-subject means ($n=9$ and $n=8$). Parameters were set on separate development subjects, except that three test subjects informed the multi-anchor candidate rule (see Ablation); for real motion, the TV weight ($\lambda=3\times10^{-3}$) was chosen on the motion-free scans by k-space cross-validation.

\begin{table}[t]
\centering
\caption{Quantitative results. (a) Simulated moderate motion (mean, 9 subjects). (b) Real motion (mean $\pm$ SD, 26 scans). (c) Ablation of EventMC-S (mean $\Delta$PSNR, 36 R1/R2 moderate and M cases); $-$/$+$: removing/adding a component. Event refs: event-guided references; Disconnection: no smoothing across events.}
\label{tab:main}
\footnotesize
\setlength{\tabcolsep}{3.0pt}
\textit{(a) CC359, simulated moderate motion}\\[1pt]
\begin{tabular}{lcccccc}
\hline
 & \multicolumn{2}{c}{R1} & \multicolumn{2}{c}{R2} & \multicolumn{2}{c}{R4}\\
Method & PSNR & SSIM & PSNR & SSIM & PSNR & SSIM\\
\hline
Uncorrected & 26.4 & .731 & 26.0 & .708 & 25.8 & .691\\
Joint-AltOpt & 31.6 & \underline{.893} & \underline{30.7} & \underline{.870} & 30.3 & .858\\
MotionTTT & 27.4 & .751 & 27.2 & .705 & 31.6 & \underline{.897}\\
MotionTTT+TV & 30.8 & .840 & 29.9 & .835 & \textbf{32.6} & \textbf{.913}\\
EventMC-S & \underline{32.6} & .884 & \textbf{33.6} & \textbf{.926} & 31.6 & .881\\
EventMC & \textbf{34.8} & \textbf{.944} & \textbf{33.6} & \textbf{.926} & \underline{32.1} & .891\\
\hline
\end{tabular}\\[4pt]
\begin{minipage}[t]{0.57\columnwidth}
\centering
\textit{(b) PMoC3D, real motion}\\[1pt]
\setlength{\tabcolsep}{2pt}
\begin{tabular}{lcc}
\hline
Method & PSNR & SSIM\\
\hline
Uncorrected (L1) & 30.7$\pm$2.0 & .931$\pm$.014\\
Stacked U-Net & 31.1$\pm$1.7 & .944$\pm$.012\\
AltOpt & 32.5$\pm$1.9 & \underline{.948}$\pm$.012\\
MotionTTT & \underline{32.6}$\pm$1.9 & \textbf{.950}$\pm$.012\\
EventMC & \textbf{32.7}$\pm$1.9 & \textbf{.950}$\pm$.012\\
\hline
\end{tabular}
\end{minipage}\hfill
\begin{minipage}[t]{0.42\columnwidth}
\centering
\textit{(c) Ablation}\\[1pt]
\setlength{\tabcolsep}{2pt}
\begin{tabular}{lc}
\hline
Change & $\Delta$PSNR\\
\hline
$-$Event refs & $-2.1$\\
$-$Disconnection & $-0.7$\\
$-$Stage A & $-6.6$\\
$-$Coarse-to-fine & $-0.3$\\
$+$Multi-anchor & $+0.5$\\
\hline
\end{tabular}
\end{minipage}
\end{table}

\subsection{Results and discussion}

\textbf{Simulated motion.} EventMC gives the best image quality at R1 and R2 under moderate motion (Table~\ref{tab:main}a, Fig.~\ref{fig:cases}a), improving PSNR and SSIM over MotionTTT and MotionTTT+TV consistently across subjects ($p\le0.012$). Since MotionTTT+TV shares our final TV reconstruction, the mean PSNR gain of up to 4.0~dB over it comes from better motion estimates; the same holds under mid-scan (M) motion (36.0/34.7 versus 32.3/32.7~dB at R1/R2, not shown in Table~\ref{tab:main}). Indeed, EventMC estimates motion more accurately than MotionTTT at R1 and R2 ($p\le0.008$). The TV prior alone is not enough: Joint-AltOpt, which uses it in plain alternation, has lower PSNR than EventMC in 53 of the 54 moderate and M cases. At R4, the sampling MotionTTT was trained for, both advantages disappear (see Limitations); MotionTTT+TV's small edge in image quality under moderate motion is not significant. Under drift alone, EventMC-S degrades no case and detects no event. In Fig.~\ref{fig:traj}, EventMC generally follows the true motion, whereas MotionTTT shows large errors in $t_y$ and $r_x$ and a false movement at about 1~min; neither method resolves the short excursion within a segment near 7.5~min (see Limitations).

\textbf{Real motion.} On PMoC3D, EventMC reaches mean image quality similar to MotionTTT and AltOpt without any training and beats Stacked U-Net on every scan ($p=0.008$; Table~\ref{tab:main}b, Fig.~\ref{fig:cases}b). It is better than MotionTTT in most scans, but not significantly; its largest losses occur in scans with frequent movements (see Limitations).

\textbf{Cost.} EventMC is also cheaper than MotionTTT. On one H100, a single run (EventMC-S) is 2.7--7.5 times faster and needs less than half the GPU memory, taking a median of under 7~min per case excluding data loading; the full EventMC takes under 12~min end to end.

\textbf{Ablation.} Removing any tested component of EventMC-S lowers mean PSNR over the 36 R1/R2 cases, most of all Stage A (Table~\ref{tab:main}c). Adding the multi-anchor extension (EventMC) fixes several failed single runs at little cost, mainly in the three test subjects that informed the candidate rule. On the remaining six subjects, EventMC retains higher mean PSNR than MotionTTT+TV at R1 and R2, although neither comparison is significant after Holm correction.

\textbf{Limitations.} First, EventMC loses its advantage at high acceleration: it no longer outperforms MotionTTT+TV at R4 and is only similar to MotionTTT and AltOpt on PMoC3D ($R\approx4.9$). A likely cause is that TV cannot reconstruct reliable references from heavily undersampled data; learned priors such as diffusion models~\cite{jalal2021,chung2022,jiang2024fast} could help. Second, EventMC assumes that head motion is mostly slow drift with occasional abrupt movements, as typically reported~\cite{zaitsev2015,qin2009}: a movement within a segment gets one compromise pose (e.g., near 7.5~min in Fig.~\ref{fig:traj}), and frequent movements leave no long stable interval for the anchor, as in its worst PMoC3D scans. Third, EventMC requires each segment to sample both low and high spatial frequencies, as in interleaved acquisition~\cite{disorder}.
\section{Conclusion}

We presented EventMC, a training-free motion correction method for interleaved 3D brain MRI that estimates motion against event-guided references and refines it from low to high spatial frequencies. In simulation, it outperforms MotionTTT at full and two-fold sampling, even with the same final TV reconstruction, and a single run needs a fraction of MotionTTT's runtime and memory. On in vivo motion-corrupted scans ($R\approx4.9$), its mean image quality is similar to MotionTTT's without any training. Future work includes further validation on independent cohorts and combining event-guided references with learned priors, such as fast diffusion models~\cite{jiang2024fast}, for high acceleration.

\section{Compliance with ethical standards}
\label{sec:ethics}

This research study was conducted retrospectively using human subject data made available in open access by CC359~\cite{cc359} and PMoC3D~\cite{pmoc3d}. Ethical approval was not required as confirmed by the license attached with the open access data.

\section{Acknowledgments}
\label{sec:acknowledgments}

WJ acknowledges support from the Dr F \& Mrs ME Zaccari Scholarship. HS acknowledges support from the Australian Research Council (DE210101297, DP230101628). The authors declare no conflicts of interest. Claude (Anthropic) assisted with text editing and code; the authors checked all methods, results and text.

 \clearpage
\appendix
\counterwithin{equation}{section}
\counterwithin{table}{section}
\counterwithin{figure}{section}
\setcounter{dbltopnumber}{3}
\renewcommand{\dblfloatpagefraction}{0.6}
\setlist[itemize]{nosep,leftmargin=13pt,topsep=2pt}
\setlist[enumerate]{nosep,leftmargin=15pt,topsep=2pt}
\renewcommand{\arraystretch}{1.16}
\setlength{\tabcolsep}{5pt}
\newcommand{\relatedto}[1]{\noindent{\small\itshape Related to #1.}\par\smallskip}
\makeatletter
\setlength{\@fptop}{0pt}
\setlength{\@dblfptop}{0pt}
\setlength{\@fpsep}{12pt plus 2pt}
\setlength{\@dblfpsep}{12pt plus 2pt}
\makeatother

\begin{center}\textbf{APPENDIX}\end{center}

This appendix describes how EventMC is implemented (Appendix~\ref{app:method}), how the experiments are set up and evaluated (Appendix~\ref{app:setup}), and what the additional results show (Appendix~\ref{app:results}). References without an appendix letter refer to the main text.\par\noindent Code: \href{https://github.com/bayesbreeze/EventMC}{\texttt{github.com/bayesbreeze/EventMC}}.

\section{Method details}
\label{app:method}

A pre-run selects the anchor. Stage A then builds event-guided references and estimates motion; Stage B refines the estimates using all data. The details below follow this order, with shared operators defined first.

\subsection{Operators and the two updates}
\label{app:ops}

\relatedto{Section~2 and the definitions of $\ImgEst$ and $\MotEst$ in Section~3}

\textbf{Poses.} A pose $\boldsymbol{\theta}_s=(\mathbf{r}_s,\mathbf{t}_s)$ moves the object as $\mathbf{u}\mapsto R(\mathbf{r}_s)(\mathbf{u}-\mathbf{c})+\mathbf{c}+\mathbf{t}_s$. Here $\mathbf{c}$ is the FOV centre and $R(\mathbf{r})$ is the rotation with rotation vector $\mathbf{r}$ (in degrees).

\textbf{Rigid transform.} $\mathcal{T}(\boldsymbol{\theta}_s)$ is implemented with Fourier operations. The rotation is factored into rotations about the three array axes, and each of these is applied as three one-dimensional Fourier shears~\cite{unser1995}. The translation is applied as a phase ramp in k-space, with periodic boundaries. Gradients with respect to the pose are obtained by automatic differentiation through this discrete operator.

\textbf{Coil sensitivities.} $S_c$ is fixed to the scanner.
\begin{itemize}
\item \emph{CC359:} $S_c$ is estimated from the motion-corrupted data $\mathbf{y}$. We take the central $24^3$ k-space samples with a 3D Hann window, transform them to image space and divide by their root-sum-of-squares over coils (clamped below at $10^{-3}$ of its maximum).
\item \emph{PMoC3D:} we use the maps supplied with the dataset, cropped to the central 256 image-space voxels along the readout and normalised by their root-sum-of-squares (clamped below at $0.1$ of its maximum).
\end{itemize}

\textbf{Uncorrected image.} The image $\mathbf{x}_0$ is the zero-filled, sensitivity-weighted coil combination
\begin{equation}
\mathbf{x}_0=\frac{\sum_c \bar{S}_c\,\mathcal{F}^{-1}\mathbf{y}_c}{\max\bigl(\sum_c \lvert S_c\rvert^2,\,10^{-6}\bigr)}.
\label{app:eq-x0}
\end{equation}

\textbf{Total variation.} The image regulariser is
\begin{equation}
\operatorname{TV}(\mathbf{x})=\sum_{\mathbf{i}}\Bigl(\sum_{d=1}^{3}\lvert D_d\mathbf{x}(\mathbf{i})\rvert^2+\varepsilon\Bigr)^{1/2},\quad \varepsilon=10^{-6},
\label{app:eq-tv}
\end{equation}
where $D_1,D_2,D_3$ are forward differences along the three axes and the sum runs over voxels whose three forward neighbours lie inside the grid. The TV weight is $\lambda=10^{-3}$ on CC359 and $\lambda=3\times10^{-3}$ on PMoC3D (Appendix~\ref{app:pmoc}). It refers to normalised intensities:
\begin{itemize}
\item \emph{CC359:} the 99.5th percentile of the motion-free image magnitude is 1. The released code uses the uncorrected image instead.
\item \emph{PMoC3D:} the 99th percentile of the uncorrected image is 1, taken over voxels above 8\% of its maximum. The scale comes from the subject's motion-free scan.
\end{itemize}

\textbf{Optimisation.} Both updates use L-BFGS~\cite{lbfgs} with a strong Wolfe line search. Image updates optimise the real and imaginary parts of $\mathbf{x}$ with history size 8 and a budget of 15 iterations in Stages A/B or 30 in the final reconstruction. Motion updates use history size 10 and 30 iterations per segment and starting pose. Each call creates a new optimiser: a warm start reuses the estimate, not the optimiser history. Motion is fitted one segment at a time, with the anchor fixed throughout.

\textbf{Central band.} $\mathcal{P}_f$ keeps the samples with $\lvert k_i\rvert\le fN_i/2$ along each k-space axis $i$, where $N_i$ is the grid size.

\subsection{Pre-run, stable interval and within-segment check}
\label{app:prerun}

\relatedto{Section~\ref{sec:stage-a} (\emph{Pre-run}; \emph{Reference segments})}

\textbf{Notation.} The events $\mathcal{J}$ divide the segments into intervals $\mathcal{I}_0,\ldots,\mathcal{I}_M$. The segments next to an event are
\begin{equation}
\mathcal{B}(\mathcal{J})=\bigcup_{j\in\mathcal{J}}\lbrace j,\,j+1\rbrace .
\label{app:eq-adjacent}
\end{equation}

\textbf{Pre-run.}
\begin{enumerate}
\item Run one Stage A round from $\Theta=\mathbf{0}$ with the temporary anchor $\alpha_0=\lfloor N_s/2\rfloor=6$. This is Round~1 below, with $\mathcal{A}$ containing all segments. Event detection gives $\mathcal{J}_{\mathrm{pre}}$.
\item On CC359, the within-segment check described below flags a set $\mathcal{G}$ of segments. On PMoC3D the check is skipped and $\mathcal{G}=\emptyset$, because each segment has only four shots.
\item Reduce each interval to its remaining segments, $\tilde{\mathcal{I}}_m=\mathcal{I}_m\setminus\bigl(\mathcal{B}(\mathcal{J}_{\mathrm{pre}})\cup\mathcal{G}\bigr)$.
\item Choose the anchor:
  \begin{itemize}
  \item $\alpha$ is the remaining segment at position $\lfloor n/2\rfloor$ (counting from 0) of the largest $\tilde{\mathcal{I}}_m$, where $n=\lvert\tilde{\mathcal{I}}_m\rvert$. Ties go to the earlier interval.
  \item $\alpha_2$ is chosen by the same rule from the second-largest non-empty $\tilde{\mathcal{I}}_m$, if one exists.
  \item If all $\tilde{\mathcal{I}}_m$ are empty, $\alpha=\alpha_0$.
  \end{itemize}
\end{enumerate}

\textbf{Stable interval.} In Rounds 2--5 of Stage A, event-adjacent segments are excluded except for the anchor:
\begin{equation}
\mathcal{A}=\mathcal{I}(\alpha)\setminus\bigl(\mathcal{B}(\mathcal{J})\setminus\lbrace\alpha\rbrace\bigr),
\label{app:eq-stable}
\end{equation}
where $\mathcal{I}(\alpha)$ is the interval that contains $\alpha$ under the current events. The anchor is always kept, so $\mathcal{A}$ is never empty, with no minimum-length fallback. The within-segment check affects anchor selection only; it is not repeated during Stage A.

\textbf{Within-segment check (CC359).}
\begin{enumerate}
\item \emph{Slices.} Transform the data to hybrid space with an inverse FFT along the readout. Keep 24 evenly spaced readout positions among those whose uncorrected-image energy exceeds 5\% of the maximum. At zero motion, each slice is a 2D multi-coil problem.
\item \emph{Sub-windows.} Split the shots of each segment, in acquisition order, into three sub-windows. With 13 segments this gives 39 sub-windows.
\item \emph{Prediction error.} For each sub-window $w$, reconstruct the slices from all other non-excluded sub-windows by 2D TV, using 40 L-BFGS iterations and the same $\lambda$. Then compute $e_w$, the squared prediction error divided by the observed energy, both summed over the selected slices.
\item \emph{Robust score.} Compare $\log e_w$ with non-excluded sub-windows at the same within-segment position (first, second or third), whose segments are at least two segments away:
\begin{equation}
z_w=\frac{\log e_w-\operatorname{med}}{\max(1.4826\,\operatorname{MAD},\,0.1)}.
\label{app:eq-z}
\end{equation}
\item \emph{Flag and repeat.} Flag $w$ if $z_w>3$. Add the flagged sub-windows to the excluded set and repeat, for at most five rounds, until no new sub-window is flagged.
\item \emph{Result.} $\mathcal{G}$ contains every segment with a sub-window flagged in the last round.
\end{enumerate}

\subsection{Stage A: references, starting poses and initialisation}
\label{app:stagea}

\relatedto{Section~\ref{sec:stage-a} (\emph{Reference segments}; \emph{Pose fitting and events}) and Algorithm~\ref{alg:single-anchor}, lines 2--8}

\textbf{References.}
\begin{itemize}
\item \emph{Round 1:} there are four references, one per group $g\in\lbrace0,1,2,3\rbrace$, each reconstructed from all segments with $s\bmod4\ne g$. Target $s$ uses the reference of its group.
\item \emph{Rounds 2--5:} following Eq.~\eqref{eq:selection}, one reference is built from $\mathcal{A}$ for targets outside $\mathcal{A}$. Four more are built from $\mathcal{A}$ without each group, for targets inside $\mathcal{A}$. Each round therefore reconstructs at most five references.
\item Every reference is an $\ImgEst$ call with the smoothed poses of the previous round fixed (zero in Round~1).
\end{itemize}

\textbf{Image initialisation.} The following start rules specify the reference-image update in Algorithm~\ref{alg:single-anchor}:
\begin{itemize}
\item All Round~1 references start from $\mathbf{x}_0$.
\item The four group references start again from $\mathbf{x}_0$ in Round~2. In Rounds 3--5, each starts from the same group's reference of the previous round.
\item The reference from $\mathcal{A}$ starts from $\mathbf{x}_0$ in every round.
\end{itemize}

\textbf{Fitting order.} Targets are fitted in increasing order of $\lvert s-\alpha\rvert$; ties go to the smaller $s$. The neighbour closer to the anchor is $n=s+1$ if $s<\alpha$ and $n=s-1$ otherwise; for the anchor's own neighbours, $\tilde{\boldsymbol{\theta}}_{\alpha}=\mathbf{0}$.

\textbf{Starting poses.} Let $\boldsymbol{\theta}^{\mathrm{prev}}$ be the smoothed poses of the previous round. Each fit is run from three starts,
\begin{equation}
\tilde{\boldsymbol{\theta}}_n,\quad
\boldsymbol{\theta}^{\mathrm{prev}}_s,\quad
\tilde{\boldsymbol{\theta}}_n+\boldsymbol{\theta}^{\mathrm{prev}}_s-\boldsymbol{\theta}^{\mathrm{prev}}_n .
\label{app:eq-starts}
\end{equation}
Each start runs $\MotEst$ on the target's own samples against its reference. The result with the smallest $\tfrac12\Vert\mathcal{P}_{\lbrace s\rbrace}(A(\Theta)\mathbf{x}_s-\mathbf{y})\Vert_2^2$ is kept. In Round~1, $\boldsymbol{\theta}^{\mathrm{prev}}=\mathbf{0}$, so the first and third starts coincide.

\textbf{Stage B and final reconstruction.} Stage B uses one shared image, initialised at $\mathbf{x}_0$ and carried between frequency bands. Each motion fit starts once from the current pose, with no event detection or smoothing; the anchor stays fixed. After Stage B, a final 30-iteration image reconstruction uses all samples and the fixed estimated poses, starting again from $\mathbf{x}_0$.

\subsection{Event detection and smoothing}
\label{app:events}

\relatedto{Section~\ref{sec:stage-a} (\emph{Pose fitting and events})}

\textbf{Displacement.} Events are detected from the raw fits $\tilde{\Theta}$. For boundary $j\in\lbrace0,\ldots,N_s-2\rbrace$, let $(\Delta\mathbf{r}_j,\Delta\mathbf{t}_j)=\tilde{\boldsymbol{\theta}}_{j+1}-\tilde{\boldsymbol{\theta}}_j$. The root-mean-square displacement~\cite{jenkinson} of six points $\mathbf{p}_i$, placed 60~mm from the FOV centre along each axis in both directions, is
\begin{equation}
m_j=\Bigl(\frac16\sum_{i=1}^{6}\bigl\Vert\bigl(R(\Delta\mathbf{r}_j)-I\bigr)\mathbf{p}_i+\Delta\mathbf{t}_j\bigr\Vert_2^2\Bigr)^{1/2}.
\label{app:eq-rms}
\end{equation}

\textbf{Threshold.} The local noise level $\sigma_j$ is 1.4826 times the median absolute deviation of $\lbrace m_{j'}:1\le\lvert j'-j\rvert\le3\rbrace$. Boundary $j$ is an event if $m_j>\max(3\sigma_j,\,0.6\,\mathrm{mm})$. The floor is in physical units, so it is the same on the 2~mm grid of Appendix~\ref{app:multi}.

\textbf{Smoothing.} Each of the six pose components is smoothed independently. Let $\vartheta_s$ be one component of $\boldsymbol{\theta}_s$ and $\tilde{\vartheta}_s$ its raw fit. We solve
\begin{equation}
\min_{\vartheta:\,\vartheta_{\alpha}=0}\ \sum_{s\ne\alpha}\frac{(\vartheta_s-\tilde{\vartheta}_s)^2}{\sigma^2}
+\lambda_{\mathrm{sm}}\sum_{i\in\mathcal{C}}\rho_{\delta}\bigl(\vartheta_{i+1}-2\vartheta_i+\vartheta_{i-1}\bigr),
\label{app:eq-smooth}
\end{equation}
with:
\begin{itemize}
\item $\mathcal{C}=\lbrace i\in\lbrace1,\ldots,N_s-2\rbrace: i-1\notin\mathcal{J},\ i\notin\mathcal{J}\rbrace$, so every second difference that touches an event is dropped;
\item the Huber penalty $\rho_{\delta}(u)=u^2$ for $\lvert u\rvert\le\delta$ and $2\delta\lvert u\rvert-\delta^2$ otherwise;
\item $\sigma=0.3$, $\lambda_{\mathrm{sm}}=1/0.15^2$ and $\delta=0.3$, in degrees or millimetres according to the component.
\end{itemize}
The problem is solved by iteratively reweighted least squares with weights $\min(1,\delta/\lvert u\rvert)$, for at most 30 iterations. A zero second difference receives weight one. Iteration stops when successive weights agree within absolute tolerance $10^{-6}$ and relative tolerance $10^{-5}$.

\subsection{Multi-anchor candidates}
\label{app:multi}

\relatedto{Section~\ref{sec:multi-anchor}}

\begin{table}[!htb]
\centering
\caption{Multi-anchor candidates. All candidates run Stage B and the final reconstruction at full resolution.}
\label{app:tab-cands}
\footnotesize
\begin{tabular}{llc}
\toprule
Candidate & Anchor & Stage A grid\\
\midrule
(i) EventMC-S & $\alpha$ & full (1~mm on CC359)\\
(ii) & $\alpha$ & half (2~mm on CC359)\\
(iii), if $\alpha_2$ exists & $\alpha_2$ & half (2~mm on CC359)\\
\bottomrule
\end{tabular}
\end{table}

\textbf{Shared steps.} All candidates use the anchors from the same full-resolution pre-run, and all run Stage B and the final reconstruction at full resolution (Table~\ref{app:tab-cands}).

\textbf{Half resolution.} Stage A keeps the central half of each k-space axis, both data and sampling pattern, so the image grid is halved. The data are multiplied by $\sqrt8$ to keep the orthonormal FFT convention. The coil sensitivities differ by dataset:
\begin{itemize}
\item \emph{CC359:} re-estimated from the cropped data as in Appendix~\ref{app:ops};
\item \emph{PMoC3D:} the supplied maps are Fourier-cropped in the same way and normalised again.
\end{itemize}
Poses are in degrees and millimetres, so they carry over to the full grid unchanged.

\textbf{Selection.} After its final reconstruction, each candidate is scored by the full-data residual $\sum_s\Vert A_s(\boldsymbol{\theta}_s)\mathbf{x}-\mathbf{y}_s\Vert_2^2$ at full resolution, and the smallest is kept.

\subsection{Parameters}
\label{app:params}

Table~\ref{app:tab-params} collects the parameters of EventMC.

\begin{table}[!htb]
\centering
\caption{Parameters of EventMC, shared by all experiments unless stated.}
\label{app:tab-params}
\footnotesize
\begin{tabular}{@{}p{0.34\columnwidth}p{0.6\columnwidth}@{}}
\toprule
Parameter & Value\\
\midrule
Segments $N_s$ & 13\\
Reference groups & $s \bmod 4$\\
Stage A rounds & 5, after a one-round pre-run with temporary anchor $\lfloor N_s/2\rfloor=6$\\
Stage B bands $f$ & 0.25, 0.5, 1, 1\\
$\ImgEst$ iterations & 15 in Stages A and B; 30 in the final reconstruction\\
$\MotEst$ iterations & 30 per segment and start; 3 starts in Stage A, 1 in Stage B\\
TV weight $\lambda$ & $10^{-3}$ (CC359), $3\times10^{-3}$ (PMoC3D)\\
TV smoothing $\varepsilon$ & $10^{-6}$\\
Event detection & probe radius 60~mm; threshold $\max(3\sigma_j, 0.6\,\mathrm{mm})$; $\sigma_j$ from 3 boundaries on each side\\
Pose smoothing & $\sigma=0.3$, $\lambda_{\mathrm{sm}}=1/0.15^2$, $\delta=0.3$\\
Within-segment check & 24 slices; 3 sub-windows per segment; $z>3$; at most 5 rounds\\
Half-resolution grid & central half of each k-space axis\\
\bottomrule
\end{tabular}
\end{table}

\FloatBarrier
\section{Experimental setup}
\label{app:setup}

\subsection{Simulated motion on CC359}
\label{app:cc359}

\relatedto{Section~4.1 (\emph{Data})}

\textbf{Data.} We use the raw 12-coil 3D T1-weighted k-space of CC359~\cite{cc359}: a $170\times218\times256$ grid ($k_z\times k_y\times k_x$, readout along $k_x$) at 1~mm isotropic resolution, with 145 of the 170 $k_z$ planes measured (partial Fourier).

\textbf{Subjects.} The test subjects are the nine subjects of MotionTTT's test split~\cite{motionttt} other than our development subject (\texttt{e14110s3\_P59904.7}):
\begin{center}
\footnotesize\ttfamily
\begin{tabular}{@{}ll@{}}
e14120s11\_P66048.7 & e14258s3\_P76800.7\\
e14313s5\_P37888.7 & e14542s5\_P52224.7\\
e14584s5\_P30208.7 & e14691s3\_P06656.7\\
e14692s5\_P14848.7 & e15183s3\_P52224.7\\
e16673s13\_P31744.7 & \\
\end{tabular}
\end{center}
None of them was used to train or select MotionTTT's network, and our algorithm parameters were set on development data outside this list. Three test subjects (e14120s11, e14258s3 and e14691s3) were inspected when the multi-anchor candidate rule was designed; Appendix~\ref{app:ablation} reports results with them removed.

\textbf{Acquisition order.}
\begin{itemize}
\item Each scan has 218 shots of 2.3~s, and each shot acquires one $k_y$ index at all measured $k_z$.
\item Segment $s$ acquires the $k_y$ indices with $k_y\equiv5s \pmod{13}$, in increasing order, which gives 16 or 17 shots per segment.
\item At R2 and R4, each segment keeps only the points of its $k_y$ indices that lie in the sampling mask.
\end{itemize}

\textbf{Sampling.}
\begin{itemize}
\item \emph{R1:} all measured lines.
\item \emph{R2:} variable-density Poisson-disc undersampling generated with our implementation of the sigpy algorithm: $R=2.05$ relative to the full grid, a fully sampled $24\times24$ centre, and a seed fixed before the experiments.
\item \emph{R4:} MotionTTT's Poisson-disc mask ($R=4.08$, fully sampled $24\times24$ centre).
\end{itemize}

\textbf{Motion model.} Ground-truth poses are drawn per shot, so motion within a segment is present but is not modelled by the estimator. Rotations act about a pivot near the neck, so the object moves by $\mathbf{t}=(I-R)\mathbf{p}+\mathbf{d}$. The pivot $\mathbf{p}$ is the head centroid offset by $U(-10,10)$~mm left--right, $U(-18,15)$~mm anterior--posterior and 42--66~mm towards the neck. The trajectory has three parts:
\begin{itemize}
\item \emph{Drift.} With $f_k=(k/217)^{\gamma}$ and $\gamma\sim U(0.7,1.5)$, superior--inferior translation $a_{\mathrm{SI}}f_k$ and nodding rotation $a_{\mathrm{nod}}f_k$. The magnitudes are $1.5$~mm and $1.0^\circ$, each multiplied by $U(0.6,1.4)$, with random signs.
\item \emph{Physiological motion.} Sinusoids of 0.05~mm at 0.25~Hz and 1.1~Hz (superior--inferior), and $0.03^\circ$ at 0.25~Hz (nodding).
\item \emph{Abrupt events (moderate).}
  \begin{itemize}
  \item \emph{Number and size:} one to three events. Each has a main rotation axis (nod, roll or shake with probabilities 0.6, 0.15 and 0.25) with an amplitude of $2^\circ$--$5^\circ$ and random sign.
  \item \emph{Other components:} the other rotation components are drawn from $\mathcal{N}(0,(0.15\,a)^2)$, where $a$ is the main amplitude. Translations are drawn from $\mathcal{N}(0,0.3^2)$~mm left--right and anterior--posterior, and $\mathcal{N}(0,1)$~mm superior--inferior.
  \item \emph{Timing:} each event ramps over 1--3 shots. With probability 0.5 it returns after a hold of 2--20 shots. Events start at shot 16 or later, keep at least two shots apart, and avoid shots 80--112 (mid-scan).
  \end{itemize}
\end{itemize}
The three motion conditions are:
\begin{itemize}
\item \textbf{M:} a single event of the same size whose transition lies within shots 88--108 and does not return. It falls near the temporary anchor $\alpha_0$.
\item \textbf{Subthreshold:} drift ($0.15$~mm, $0.15^\circ$) and physiological motion only.
\item \textbf{Moderate:} the events described above.
\end{itemize}

\textbf{Simulation.}
\begin{itemize}
\item The motion-free complex image is moved per shot by an oversampled Fourier warp ($2\times$ oversampling, 32-voxel zero padding along the readout) and encoded with static coil maps estimated from the motion-free data.
\item Complex Gaussian noise is added once per case. Its level matches the background noise of the original coil images.
\item Each estimator receives only $\mathbf{y}$, the sampling pattern and coil maps estimated from $\mathbf{y}$. The generator and the estimator therefore differ in discretisation and coil maps.
\end{itemize}

\textbf{Cases.} Nine subjects $\times$ \{moderate, M\} $\times$ \{R1, R2, R4\} give 54 cases, plus 27 subthreshold cases. The trajectory and noise of a subject and condition are shared across R1, R2 and R4.

\subsection{Real motion on PMoC3D}
\label{app:pmoc}

\relatedto{Section~4.1 (\emph{Data}) and the choice of the TV weight}

\textbf{Data.}
\begin{itemize}
\item We use the multi-coil k-space of PMoC3D~\cite{pmoc3d} (eight subjects).
\item Each scan has 52 shots of 204 $(k_y,k_z)$ points with full readouts ($R\approx4.94$). Readout oversampling is removed by an inverse FFT along the readout, a central image-space crop to 256 voxels, and an FFT back to k-space.
\item Segments are four consecutive shots ($N_s=13$).
\item Each subject has a motion-free scan (run-00) and three or four motion scans, 27 in total. We exclude sub-05 run-04: it has 13 receiver coils, while the subject's other scans and coil maps have 14, and which coil is missing is not documented.
\end{itemize}

\textbf{TV weight.} The weight was selected on the motion-free scans only, by k-space cross-validation (Table~\ref{app:tab-lambda}):
\begin{enumerate}
\item For each subject, hold out 10\% of the sampled $(k_y,k_z)$ points (fixed seed).
\item Reconstruct from the remaining 90\% with zero motion (30 iterations, as the final reconstruction).
\item Compute the normalised error of predicting the held-out points, $\Vert A_{\mathrm{ho}}\mathbf{x}-\mathbf{y}_{\mathrm{ho}}\Vert_2^2/\Vert\mathbf{y}_{\mathrm{ho}}\Vert_2^2$.
\end{enumerate}

\begin{table}[!htb]
\centering
\caption{k-space cross-validation of the TV weight on the PMoC3D motion-free scans: median held-out error over eight subjects. Lower is better; bold marks the minimum.}
\label{app:tab-lambda}
\footnotesize
\setlength{\tabcolsep}{4pt}
\begin{tabular}{lcccccc}
\toprule
$\lambda$ & $10^{-4}$ & $3{\times}10^{-4}$ & $10^{-3}$ & $3{\times}10^{-3}$ & $10^{-2}$ & $3{\times}10^{-2}$\\
\midrule
Error & 0.1140 & 0.1009 & 0.0869 & \textbf{0.0807} & 0.0816 & 0.0941\\
\bottomrule
\end{tabular}
\end{table}

The lowest median held-out error occurs at an interior grid point, giving $\lambda=3\times10^{-3}$. Image-based metrics against the reference reconstructions were computed only as diagnostics and did not enter the selection. A further check on simulated motion was specified in advance. It would have changed the choice only if another weight improved median PSNR by more than 0.3~dB. It was planned for three subjects and completed for two, and it did not change the choice. The selected weight was fixed before the 26-scan production evaluation.

\textbf{Comparisons.} We compare with the reconstructions released with the dataset (MotionTTT, AltOpt, Stacked U-Net and uncorrected L1); none of these methods was re-run. The within-segment check is skipped (Appendix~\ref{app:prerun}).

\subsection{Baselines in simulation}
\label{app:baselines}

\relatedto{Section~4.1 (\emph{Baselines and evaluation})}

\textbf{MotionTTT}~\cite{motionttt}.
\begin{itemize}
\item We use the official code (commit 8fdc42f) and pretrained 2D U-Net (\texttt{mli-lab/}\allowbreak\texttt{Unet48-2D-CC359}). The network was trained without motion on R4 slices of 40 CC359 training subjects and stays frozen at test time.
\item We use the inter-shot settings of its simulated-data scripts, with $N_s=13$: learning rate 4.0, 130 test-time steps, data-consistency threshold 0.575 at step 71, and all motion states optimised from step 100.
\item The native output is its L1-wavelet reconstruction with data-consistency rejection.
\item Branches that use the ground truth are disabled. Only memory-related settings were changed.
\item It receives the same observations, converted to its data layout and intensity scale.
\end{itemize}

\textbf{MotionTTT+TV.} MotionTTT's segment poses are converted to our pose convention (the conversion was calibrated on the development subject). They are then reconstructed with our final reconstruction (30 iterations from $\mathbf{x}_0$, all samples).

\textbf{Forward-model difference.} MotionTTT moves the coil-weighted image $S_c\mathbf{x}$, which assumes that the coils move with the head. Our simulation and EventMC keep the coils fixed. On the development subject, the two models differ by 23\% in k-space (relative norm) for a $3^\circ$ rotation plus 2~mm translation over 12 coils. Our simulation follows the fixed-coil model, which favours EventMC in this respect.

\textbf{Joint-AltOpt}~\cite{disorder}. Alternation of Eq.~\eqref{eq:objective} on all segments from $\Theta=\mathbf{0}$, at the same anchor as EventMC-S, with three rounds at $f=0.25$, two at $f=0.5$ and two at $f=1$, followed by the final reconstruction. It uses the same operators as Stage B.

\textbf{Uncorrected.} The final reconstruction with $\Theta=\mathbf{0}$.

{\sloppy \textbf{Stacked U-Net}~\cite{almasni2022}. Author weights (\texttt{mli-lab/}\allowbreak\texttt{StackedUnet64-}\allowbreak\texttt{2D-CC359-}\allowbreak\texttt{MildSimMotion-v1.0}), applied slice by slice. Its output is rescaled to our intensity by an operator-based factor that does not use the ground truth.\par}

\subsection{Evaluation and statistics}
\label{app:eval}

\relatedto{Section~4.1 (\emph{Baselines and evaluation})}

\textbf{Image quality in simulation.}
\begin{enumerate}
\item The reference is the magnitude of the motion-free image.
\item Each method's native magnitude is rigidly aligned to the reference, because every method has its own reference frame:
  \begin{itemize}
  \item both images are smoothed with a 4~mm Gaussian;
  \item the six pose parameters are estimated with 60 L-BFGS steps and a Fourier rigid transform;
  \item the native magnitude is then resampled once.
  \end{itemize}
\item The evaluation mask is the reference head mask intersected with the valid FOV of the aligned image.
\item \emph{PSNR:} the peak is the reference maximum inside the head mask, and the mean squared error is averaged over the evaluation mask.
\item \emph{SSIM:} a 3D Gaussian window ($\sigma=1.5$, $11^3$) with $K_1=0.01$, $K_2=0.03$ and data range equal to the peak, averaged over the evaluation mask.
\item No intensity fitting is applied.
\end{enumerate}

\textbf{Pose accuracy.}
\begin{itemize}
\item Estimated segment poses are compared with the true segment-mean poses.
\item Each estimated trajectory is first right-composed with one global rigid transform, chosen to minimise the six-point displacement error over the 13 segments.
\item For each segment we compute the translation error $\Vert\hat{\mathbf{t}}_s-\mathbf{t}_s\Vert_2$ (mm) and the geodesic rotation error (degrees). We report the root mean square over segments.
\end{itemize}

\textbf{Image quality on PMoC3D.} We follow the MotionDPS protocol~\cite{motiondps} in our own implementation:
\begin{enumerate}
\item Each reconstruction is scaled by least squares to the reference, the dataset's L1 reconstruction of the subject's motion-free scan.
\item It is rigidly aligned to the reference as above.
\item A brain mask is extracted from the reference only (FastSurfer v2.5.4~\cite{fastsurfer}) and applied to both volumes.
\item Each volume is divided by its own 99.9th percentile inside the brain.
\item PSNR and SSIM are computed over the whole volume, with zeros outside the brain. The data range is the reference maximum, a per-subject constant that is the same for all methods.
\end{enumerate}

\textbf{Statistics.}
\begin{itemize}
\item \emph{Tests:} two-sided Wilcoxon signed-rank tests on per-subject means of the per-case differences ($n=9$ for CC359, $n=8$ for PMoC3D). For an exact test with $n$ nonzero paired differences, the smallest attainable $p$ is $0.0039$ for $n=9$, $0.0078$ for $n=8$ and $0.031$ for $n=6$.
\item \emph{Primary comparisons:}
  \begin{itemize}
  \item CC359: EventMC versus MotionTTT+TV in PSNR at R1 and R2, Holm-corrected over these two;
  \item PMoC3D: EventMC versus MotionTTT in PSNR, specified before the runs.
  \end{itemize}
\item \emph{Other comparisons:} exploratory and unadjusted.
\end{itemize}

\textbf{Platform.} For the 54 moderate/M simulation cases, EventMC, EventMC-S and the ablation variants ran on an H100 SXM GPU; all other simulation methods (the external baselines, Joint-AltOpt, Uncorrected and the true-pose reference) ran on an H100 PCIe. Drift-only controls use the original H100 PCIe runs. All reported timings were measured on an H100 PCIe (Appendix~\ref{app:cost}). PMoC3D baseline images were supplied with the dataset, so their reconstruction hardware is not part of this comparison.

\FloatBarrier
\section{Additional results}
\label{app:results}

\subsection{Simulated motion: all methods and both motion types}
\label{app:sim}

\relatedto{Table~\ref{tab:main}(a) and Section~4.2 (\emph{Simulated motion})}

Tables~\ref{app:tab-moderate} and~\ref{app:tab-m} give means over nine subjects. ``True segment poses'' is our final reconstruction with each segment's true mean pose. It is a reference, not a method: when a movement falls within a segment, the mean pose lies between two head positions, so this reference can score below EventMC.

\begin{table*}[!tp]
\centering
\caption{Simulated moderate motion (mean over nine subjects; PSNR in dB). Rows other than Stacked U-Net and True segment poses repeat Table~\ref{tab:main}(a). Bold: best method values, including ties at the displayed precision; the true-pose reference is excluded.}
\label{app:tab-moderate}
\small
\setlength{\tabcolsep}{3pt}
\begin{tabular*}{\textwidth}{@{\extracolsep{\fill}}lcccccc@{}}
\toprule
 & \multicolumn{2}{c}{R1} & \multicolumn{2}{c}{R2} & \multicolumn{2}{c}{R4}\\
\cmidrule(lr){2-3}\cmidrule(lr){4-5}\cmidrule(lr){6-7}
Method & PSNR $\uparrow$ & SSIM $\uparrow$ & PSNR $\uparrow$ & SSIM $\uparrow$ & PSNR $\uparrow$ & SSIM $\uparrow$\\
\midrule
Uncorrected & 26.4 & 0.731 & 26.0 & 0.708 & 25.8 & 0.691\\
Stacked U-Net & 26.5 & 0.771 & 26.3 & 0.753 & 26.3 & 0.750\\
Joint-AltOpt & 31.6 & 0.893 & 30.7 & 0.870 & 30.3 & 0.858\\
MotionTTT & 27.4 & 0.751 & 27.2 & 0.705 & 31.6 & 0.897\\
MotionTTT+TV & 30.8 & 0.840 & 29.9 & 0.835 & \textbf{32.6} & \textbf{0.913}\\
EventMC-S & 32.6 & 0.884 & \textbf{33.6} & \textbf{0.926} & 31.6 & 0.881\\
EventMC & \textbf{34.8} & \textbf{0.944} & \textbf{33.6} & \textbf{0.926} & 32.1 & 0.891\\
\midrule
True segment poses & 33.9 & 0.928 & 32.8 & 0.911 & 31.7 & 0.890\\
\bottomrule
\end{tabular*}
\end{table*}

\begin{table*}[!tp]
\centering
\caption{Simulated sustained mid-scan motion, M (mean over nine subjects; PSNR in dB). Bold: best method values, including ties at the displayed precision; the true-pose reference is excluded.}
\label{app:tab-m}
\small
\setlength{\tabcolsep}{3pt}
\begin{tabular*}{\textwidth}{@{\extracolsep{\fill}}lcccccc@{}}
\toprule
 & \multicolumn{2}{c}{R1} & \multicolumn{2}{c}{R2} & \multicolumn{2}{c}{R4}\\
\cmidrule(lr){2-3}\cmidrule(lr){4-5}\cmidrule(lr){6-7}
Method & PSNR $\uparrow$ & SSIM $\uparrow$ & PSNR $\uparrow$ & SSIM $\uparrow$ & PSNR $\uparrow$ & SSIM $\uparrow$\\
\midrule
Uncorrected & 23.9 & 0.608 & 23.7 & 0.589 & 23.6 & 0.577\\
Stacked U-Net & 23.7 & 0.645 & 23.9 & 0.641 & 24.1 & 0.653\\
Joint-AltOpt & 29.9 & 0.865 & 29.4 & 0.845 & 29.2 & 0.838\\
MotionTTT & 31.0 & 0.846 & 31.6 & 0.878 & 31.9 & 0.911\\
MotionTTT+TV & 32.3 & 0.858 & 32.7 & 0.891 & \textbf{33.3} & \textbf{0.934}\\
EventMC-S & \textbf{36.0} & \textbf{0.964} & \textbf{34.8} & \textbf{0.952} & 32.9 & 0.925\\
EventMC & \textbf{36.0} & \textbf{0.964} & 34.7 & 0.951 & \textbf{33.3} & 0.933\\
\midrule
True segment poses & 36.0 & 0.960 & 34.7 & 0.948 & 33.4 & 0.930\\
\bottomrule
\end{tabular*}
\end{table*}

\begin{figure*}[!tp]
\centering
\includegraphics[width=0.78\textwidth]{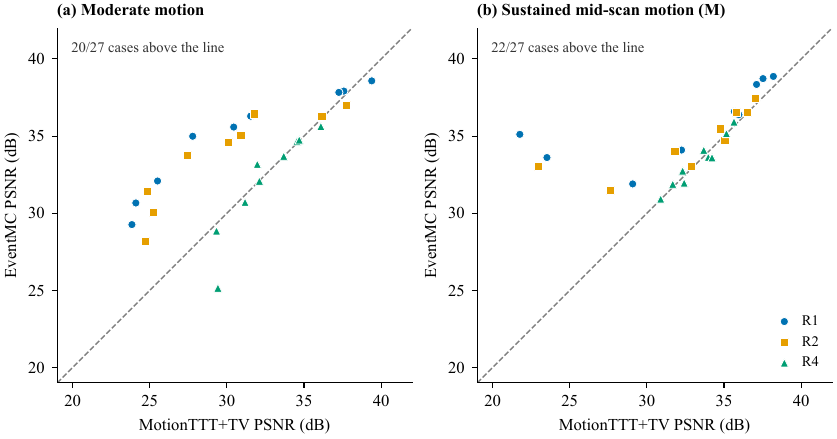}
\caption{Per-case PSNR in simulation, EventMC against MotionTTT+TV: (a) moderate and (b) M motion, 27 cases each (nine subjects at R1, R2 and R4). Points above the diagonal: EventMC is better.}
\label{app:fig-scatter}
\end{figure*}

Table~\ref{app:tab-subj} compares EventMC with each baseline at the subject level, and Fig.~\ref{app:fig-scatter} shows every case of the comparison with MotionTTT+TV, which shares our final reconstruction. The two primary comparisons, EventMC versus MotionTTT+TV in PSNR at R1 and R2, have Holm-adjusted $p=0.016$. Over all 54 cases, EventMC has higher PSNR than Joint-AltOpt in 53. Qualitative examples are collected in Appendix~\ref{app:qual}.

\begin{table*}[!tp]
\centering
\caption{Subject-level comparisons in simulation, moderate and M together. Subj.: subjects in which EventMC is better (of 9); $p$: two-sided Wilcoxon; Cases: cases in which EventMC is better (of 18).}
\label{app:tab-subj}
\small
\begin{tabular}{llccccccccc}
\toprule
 & & \multicolumn{3}{c}{R1} & \multicolumn{3}{c}{R2} & \multicolumn{3}{c}{R4}\\
\cmidrule(lr){3-5}\cmidrule(lr){6-8}\cmidrule(lr){9-11}
EventMC versus & Metric & Subj. & $p$ & Cases & Subj. & $p$ & Cases & Subj. & $p$ & Cases\\
\midrule
MotionTTT & PSNR & 9 & 0.0039 & 18 & 9 & 0.0039 & 17 & 8 & 0.020 & 16\\
 & SSIM & 9 & 0.0039 & 17 & 8 & 0.0078 & 17 & 8 & 0.13 & 16\\
MotionTTT+TV & PSNR & 8 & 0.0078 & 17 & 8 & 0.012 & 15 & 4 & 0.57 & 10\\
 & SSIM & 8 & 0.0078 & 17 & 8 & 0.012 & 15 & 4 & 0.43 & 9\\
Joint-AltOpt & PSNR & 9 & 0.0039 & 18 & 9 & 0.0039 & 18 & 9 & 0.0039 & 17\\
 & SSIM & 9 & 0.0039 & 18 & 9 & 0.0039 & 18 & 9 & 0.0039 & 17\\
\bottomrule
\end{tabular}
\end{table*}

\subsection{Pose accuracy}
\label{app:pose}

\relatedto{Section~4.2 (\emph{Simulated motion})}

Table~\ref{app:tab-pose} gives the pose errors, defined in Appendix~\ref{app:eval}. Fig.~\ref{app:fig-traj} shows all six pose components for the case of Fig.~\ref{fig:traj}.

\begin{table*}[!tp]
\centering
\caption{Pose error: median per-case RMSE over 18 cases per sampling level (moderate and M). Lower is better; bold marks the minimum at the displayed precision. The last column gives exploratory, unadjusted subject-level Wilcoxon p-values for EventMC versus MotionTTT ($n=9$).}
\label{app:tab-pose}
\small
\begin{tabular*}{\textwidth}{@{\extracolsep{\fill}}llccccc@{}}
\toprule
Sampling & Error $\downarrow$ & Joint-AltOpt & MotionTTT & EventMC-S & EventMC & $p$\\
\midrule
R1 & Translation (mm) & 0.57 & 0.52 & 0.30 & \textbf{0.29} & 0.0078\\
   & Rotation (deg) & 0.54 & 0.50 & 0.31 & \textbf{0.26} & 0.0078\\
\addlinespace[2pt]
R2 & Translation (mm) & 0.57 & 0.47 & 0.27 & \textbf{0.25} & 0.0039\\
   & Rotation (deg) & 0.55 & 0.56 & \textbf{0.26} & \textbf{0.26} & 0.0078\\
\addlinespace[2pt]
R4 & Translation (mm) & 0.56 & 0.30 & 0.31 & \textbf{0.19} & 0.43\\
   & Rotation (deg) & 0.51 & 0.30 & 0.33 & \textbf{0.24} & 0.50\\
\bottomrule
\end{tabular*}
\end{table*}

\begin{figure*}[!tp]
\centering
\includegraphics[width=\textwidth]{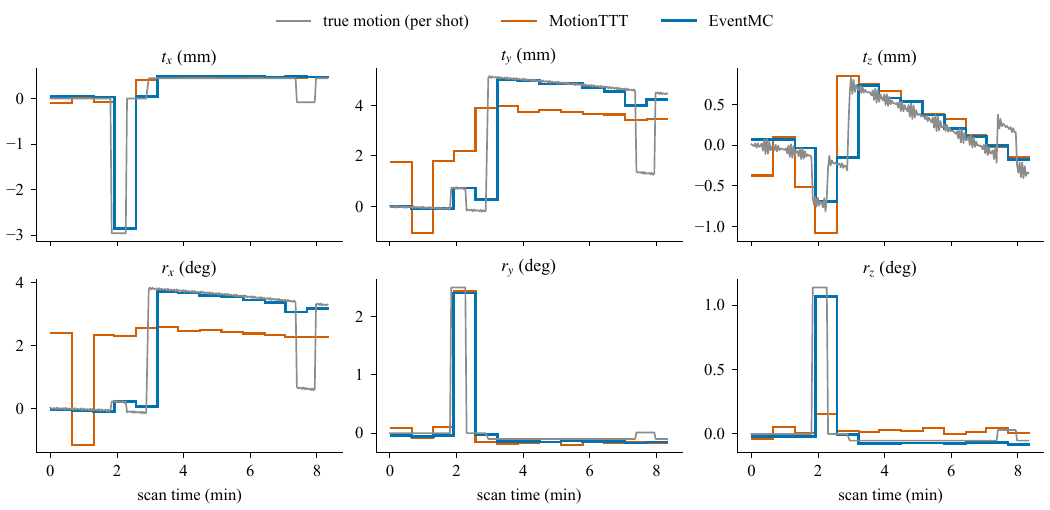}
\caption{All six pose components for the case of Fig.~\ref{fig:traj} (CC359, moderate motion, R1). The true motion is shown per shot; each method estimates one pose per segment. Estimates are shown after removing the global rigid offset (Appendix~\ref{app:eval}).}
\label{app:fig-traj}
\end{figure*}

\subsection{Drift-only control}
\label{app:drift}

\relatedto{Section~4.2 (\emph{Simulated motion})}

On the 27 subthreshold cases, EventMC-S detects no event in the pre-run or in any Stage A round. Table~\ref{app:tab-drift} compares it with the uncorrected reconstruction.

\begin{table}[!htb]
\centering
\caption{Drift-only control: EventMC-S minus the uncorrected reconstruction on the 27 subthreshold cases ($\Delta$PSNR in dB; counts out of nine cases).}
\label{app:tab-drift}
\footnotesize
\setlength{\tabcolsep}{4pt}
\begin{tabular}{lcccc}
\toprule
 & Median & Smallest & Cases with & Cases with\\
Sampling & $\Delta$PSNR & $\Delta$PSNR & lower PSNR & lower SSIM\\
\midrule
R1 & $+0.44$ & $+0.22$ & 0 & 0\\
R2 & $+0.28$ & $+0.15$ & 0 & 0\\
R4 & $+0.17$ & $+0.12$ & 0 & 0\\
\bottomrule
\end{tabular}
\end{table}

\FloatBarrier
\subsection{Ablation and the multi-anchor extension}
\label{app:ablation}

\relatedto{Table~\ref{tab:main}(c) and Section~4.2 (\emph{Ablation})}

\textbf{Variants.} Each variant changes one component of EventMC-S (Table~\ref{app:tab-variants}). It keeps the anchor chosen by EventMC-S's pre-run and ends with the final reconstruction. Differences are relative to EventMC-S run on the same platform.

\begin{table}[!htb]
\centering
\caption{Ablation variants of EventMC-S.}
\label{app:tab-variants}
\footnotesize
\begin{tabular}{@{}p{0.27\columnwidth}p{0.67\columnwidth}@{}}
\toprule
Variant & What changes\\
\midrule
$-$Event refs & All five Stage A rounds use the Round~1 references (all segments except the target's group). Events are still detected and still split the smoothing.\\
$-$Disconnection & Smoothing also spans events (all second differences are kept). Events still select the references.\\
$-$Stage A & Stage B starts from $\Theta=\mathbf{0}$.\\
$-$Coarse-to-fine & Stage B is replaced by four rounds at $f=1$, starting from the Stage A poses.\\
$+$Multi-anchor & EventMC (Appendix~\ref{app:multi}).\\
\bottomrule
\end{tabular}
\end{table}

\textbf{Results.} Removing Stage A produces the largest average loss (Table~\ref{app:tab-ablation}). Event-guided references are particularly important under M motion, while disconnection has a larger median effect under moderate motion.

\begin{table*}[!tp]
\centering
\caption{Ablation on the 36 R1 and R2 cases, relative to EventMC-S on the same platform ($\Delta$PSNR in dB). Subject-level tests are exploratory and unadjusted; ties are omitted from better/worse counts.}
\label{app:tab-ablation}
\small
\begin{tabular}{lccccc}
\toprule
 & Mean & Median $\Delta$PSNR & Cases & \multicolumn{2}{c}{Subjects better / worse ($p$)}\\
\cmidrule(lr){5-6}
Variant & $\Delta$PSNR & moderate / M & below $-1$~dB & R1 & R2\\
\midrule
$-$Event refs & $-2.1$ & $-0.04$ / $-4.11$ & 16 & 2 / 7 (0.039) & 0 / 9 (0.0039)\\
$-$Disconnection & $-0.7$ & $-1.07$ / $-0.48$ & 13 & 3 / 6 (0.43) & 0 / 9 (0.0039)\\
$-$Stage A & $-6.6$ & $-4.88$ / $-8.56$ & 35 & 0 / 9 (0.0039) & 0 / 9 (0.0039)\\
$-$Coarse-to-fine & $-0.3$ & $0.00$ / $0.00$ & 4 & 3 / 6 (0.30) & 6 / 3 (0.91)\\
$+$Multi-anchor & $+0.5$ & $0.00$ / $0.00$ & 0 & 4 / 1 (0.125) & 2 / 1 (1.0)\\
\bottomrule
\end{tabular}
\end{table*}

\begin{figure}[!htb]
\centering
\includegraphics[width=0.85\columnwidth]{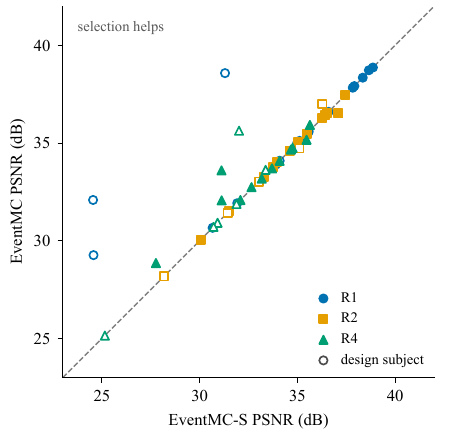}
\caption{Multi-anchor selection on the 54 moderate and M cases: PSNR of the selected candidate (EventMC) against candidate~(i) (EventMC-S). Points on the diagonal: the rule kept candidate~(i) or changed little. Hollow markers: the three subjects inspected when the rule was designed. Here PSNR is computed within the intersection of the candidates' valid FOVs, so values can differ slightly from Table~\ref{app:tab-moderate}.}
\label{app:fig-sel}
\end{figure}

\textbf{Multi-anchor selection.} Fig.~\ref{app:fig-sel} shows all 54 moderate and M cases (R1, R2 and R4). Candidate PSNRs are evaluated on the common intersection of their valid FOVs.
\begin{itemize}
\item \emph{Which candidate wins:} the residual rule keeps candidate (i), EventMC-S, in 36 cases, candidate (ii) in 8 and candidate (iii) in 10.
\item \emph{Agreement with image quality:} the selected candidate is within 0.1 dB of the highest candidate PSNR in 50 of 54 cases.
\item \emph{Gains and losses relative to EventMC-S:} PSNR rises by more than 1~dB in six cases (at most 7.5~dB) and falls by more than 0.1~dB in three (at most 0.58~dB).
\item \emph{Design subjects:} four of the six large gains are in the three subjects inspected when the rule was designed.
\end{itemize}

\textbf{Sensitivity to the design subjects.} Table~\ref{app:tab-sens} repeats the primary comparison without the three design subjects. EventMC remains better in all six subjects at R1 and R2, but neither comparison is significant after Holm correction ($p=0.063$). With six nonzero paired differences, the minimum exact two-sided p-value is $0.03125$; correcting these two comparisons yields a minimum of $0.0625$.

\begin{table*}[!tp]
\centering
\caption{Sensitivity to the three design subjects, moderate and M together: mean PSNR (dB) of EventMC versus MotionTTT+TV, and subjects in which EventMC is better. Bold marks the higher mean within each pair; $p$ is raw, followed by its Holm-adjusted value.}
\label{app:tab-sens}
\small
\begin{tabular}{lcccc}
\toprule
 & \multicolumn{2}{c}{R1} & \multicolumn{2}{c}{R2}\\
\cmidrule(lr){2-3}\cmidrule(lr){4-5}
Cohort & Mean PSNR & Subjects better ($p$, Holm) & Mean PSNR & Subjects better ($p$, Holm)\\
\midrule
All nine subjects & \textbf{35.4} vs 31.6 & 8/9 (0.0078, 0.016) & \textbf{34.2} vs 31.3 & 8/9 (0.012, 0.016)\\
Without the three design subjects & \textbf{36.2} vs 31.9 & 6/6 (0.031, 0.063) & \textbf{34.9} vs 31.7 & 6/6 (0.031, 0.063)\\
\bottomrule
\end{tabular}
\end{table*}

\subsection{Computational cost}
\label{app:cost}

\relatedto{Section~4.2 (\emph{Cost})}

\textbf{Setup.}
\begin{itemize}
\item All timings are on one H100 PCIe GPU, exclusively allocated.
\item Simulation, data loading and evaluation are excluded. The stages of each case are summed before the median is taken.
\item MotionTTT, Joint-AltOpt and EventMC-S were timed on all 18 cases at each sampling level. The full EventMC was timed end to end on two cases per level with a separate driver.
\end{itemize}

Peak memory is the range across cases. The end-to-end multi-candidate run uses a separate driver, whose live tensors differ from the single-candidate driver; the lower recorded peak does not imply that adding candidates reduces memory.

\begin{figure*}[!tp]
\centering
\includegraphics[width=\textwidth]{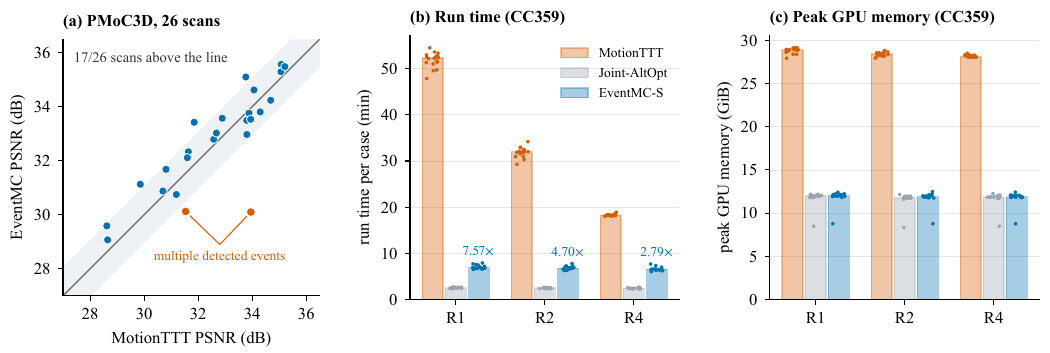}
\caption{(a) Per-scan PSNR on PMoC3D, EventMC against MotionTTT; the band marks $\pm1$~dB, and the orange points are the two scans with the largest losses (Appendix~\ref{app:real}). (b,~c) Run time and peak GPU memory per case on CC359 (one H100 PCIe). Bars: medians; dots: cases; labels: ratio of the median run times of MotionTTT and EventMC-S.}
\label{app:fig-realcost}
\end{figure*}

\textbf{Results.} Table~\ref{app:tab-cost} and Fig.~\ref{app:fig-realcost}b,c give the run times and memory. The ratios of median run times relative to MotionTTT are 7.57 (R1), 4.70 (R2) and 2.79 (R4) for EventMC-S, and 4.50, 2.86 and 1.67 for EventMC.

\begin{table*}[!tp]
\centering
\caption{Run time per case in minutes (median, with range) and peak GPU memory range on CC359, one H100 PCIe. Bold marks the lowest median run time. Memory ranges are not ranked across the different drivers.}
\label{app:tab-cost}
\small
\begin{tabular}{lcccc}
\toprule
Method & R1 & R2 & R4 & Peak GPU memory (GiB)\\
\midrule
MotionTTT & 52.3 (47.8--54.5) & 32.0 (29.3--34.2) & 18.3 (18.0--18.9) & 27.9--29.1\\
Joint-AltOpt & \textbf{2.5} (2.4--2.8) & \textbf{2.5} (2.3--2.7) & \textbf{2.5} (2.3--2.7) & 8.3--12.3\\
EventMC-S & 6.9 (6.5--7.9) & 6.8 (6.3--7.8) & 6.6 (6.0--7.7) & 8.8--12.5\\
EventMC (2 cases per level) & 11.6 (11.4--11.8) & 11.2 (11.1--11.3) & 10.9 (10.9--11.0) & 8.7--9.3\\
\bottomrule
\end{tabular}
\end{table*}

\subsection{Real motion: per-subject results}
\label{app:real}

\relatedto{Table~\ref{tab:main}(b) and Section~4.2 (\emph{Real motion}; \emph{Limitations})}

Table~\ref{app:tab-real} compares EventMC with each method, and Fig.~\ref{app:fig-realcost}a shows every scan.

\begin{table}[!htb]
\centering
\caption{Real motion: EventMC versus each method, PSNR, 26 scans of eight subjects. $\Delta$: median per-scan PSNR difference (dB).}
\label{app:tab-real}
\footnotesize
\setlength{\tabcolsep}{4pt}
\begin{tabular}{lccc}
\toprule
EventMC versus & Subjects better ($p$) & Scans better & $\Delta$\\
\midrule
MotionTTT & 5/8 (0.64) & 17/26 & $+0.31$\\
AltOpt & 5/8 (0.46) & 17/26 & $+0.41$\\
Stacked U-Net & 8/8 (0.0078) & 26/26 & $+1.54$\\
Uncorrected (L1) & 8/8 (0.0078) & 26/26 & $+1.92$\\
\bottomrule
\end{tabular}
\end{table}

\textbf{Primary comparison.} For PSNR, EventMC improves on MotionTTT in 17 of 26 scans and 5 of 8 subjects, with a median scan-level gain of 0.31~dB ($p=0.64$ at the subject level). This meets our pre-specified descriptive criterion for similar performance: a nonsignificant difference and a median scan-level difference within $\pm0.5$~dB. It is not a formal equivalence test. SSIM gives the same subject-level count and p-value, with improvements in 18 of 26 scans.

\textbf{Without motion correction.} Our TV reconstruction without motion correction (not in Table~\ref{tab:main}(b)) reaches $31.1\pm2.1$~dB and $0.935\pm0.015$. EventMC improves on it in 25 of 26 scans; the exception is sub-07 run-03, at $-0.14$~dB.

\textbf{Per subject.} Table~\ref{app:tab-realsubj} gives the per-subject differences from MotionTTT.

\begin{table}[!htb]
\centering
\caption{Real motion: per-subject mean PSNR difference, EventMC minus MotionTTT (dB).}
\label{app:tab-realsubj}
\footnotesize
\setlength{\tabcolsep}{3pt}
\begin{tabular}{cccccccc}
\toprule
sub-01 & sub-02 & sub-03 & sub-04 & sub-05 & sub-06 & sub-07 & sub-08\\
\midrule
$+0.68$ & $-0.79$ & $+0.19$ & $+0.42$ & $+0.52$ & $+0.89$ & $-0.25$ & $-0.61$\\
\bottomrule
\end{tabular}
\end{table}

\textbf{Largest differences from MotionTTT.} Fig.~\ref{app:fig-pmoc} shows the scans with the largest gain and the largest loss.
\begin{itemize}
\item \emph{Largest gains:} sub-08 run-02 ($+1.57$~dB; Fig.~\ref{fig:cases}b), sub-05 run-03 ($+1.33$~dB) and sub-06 run-01 ($+1.27$~dB).
\item \emph{Largest losses:} sub-08 run-01 ($-3.85$~dB) and sub-02 run-01 ($-1.42$~dB).
  \begin{itemize}
  \item Their pre-runs detected four and seven events. The largest remaining interval lay at the end of the scan, which put the anchor in segment 11 and segment 12.
  \item The half-resolution candidates did not improve either scan.
  \end{itemize}
\end{itemize}

\textbf{Multi-anchor on real data.} On PMoC3D, the residual rule chose a half-resolution candidate in one scan (sub-01 run-03, 0.03~dB below EventMC-S). Otherwise EventMC and EventMC-S are identical.

\subsection{Qualitative examples}
\label{app:qual}

\relatedto{Fig.~\ref{fig:cases} and Section~4.2}

\begin{figure*}[p]
\centering
\includegraphics[width=\textwidth]{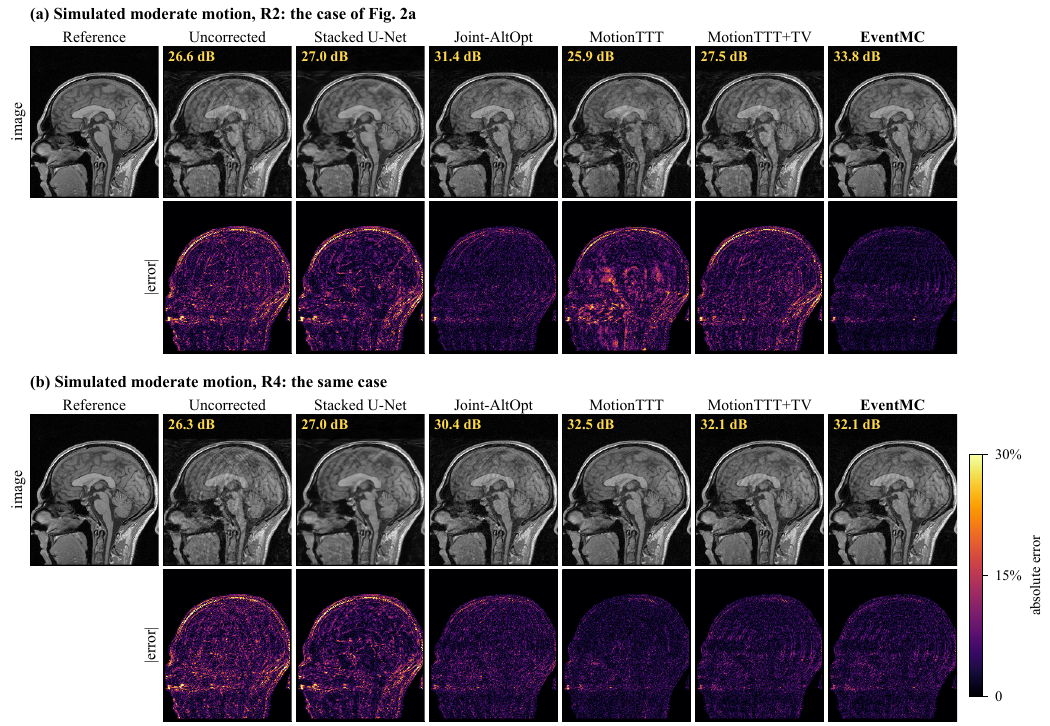}
\caption{Central sagittal slices of the case of Fig.~\ref{fig:cases}a (CC359, moderate motion) at (a) R2 and (b) R4, with volumetric PSNR. Bottom rows: absolute error relative to the reference display range, within the head mask and each method's valid FOV.}
\label{app:fig-accel}
\end{figure*}

\begin{figure*}[p]
\centering
\includegraphics[width=\textwidth]{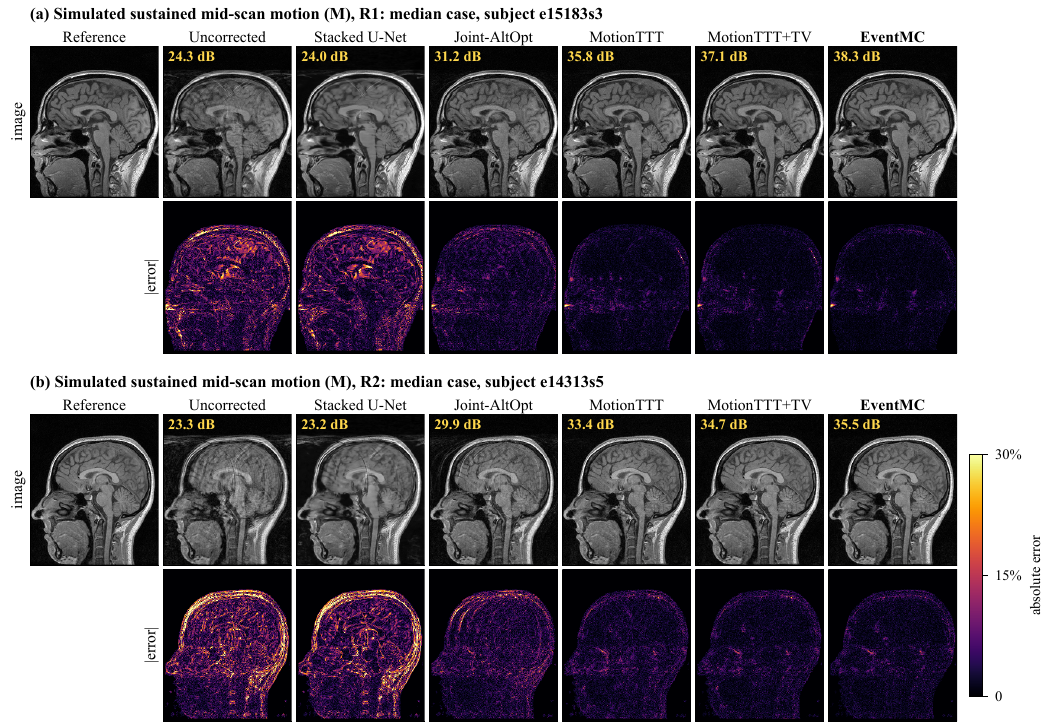}
\caption{Sustained mid-scan motion (M): central sagittal slices of the case with the median PSNR gain of EventMC over MotionTTT+TV among the nine subjects, (a) at R1 and (b) at R2. Display as in Fig.~\ref{app:fig-accel}.}
\label{app:fig-m}
\end{figure*}

\begin{figure*}[p]
\centering
\includegraphics[width=\textwidth]{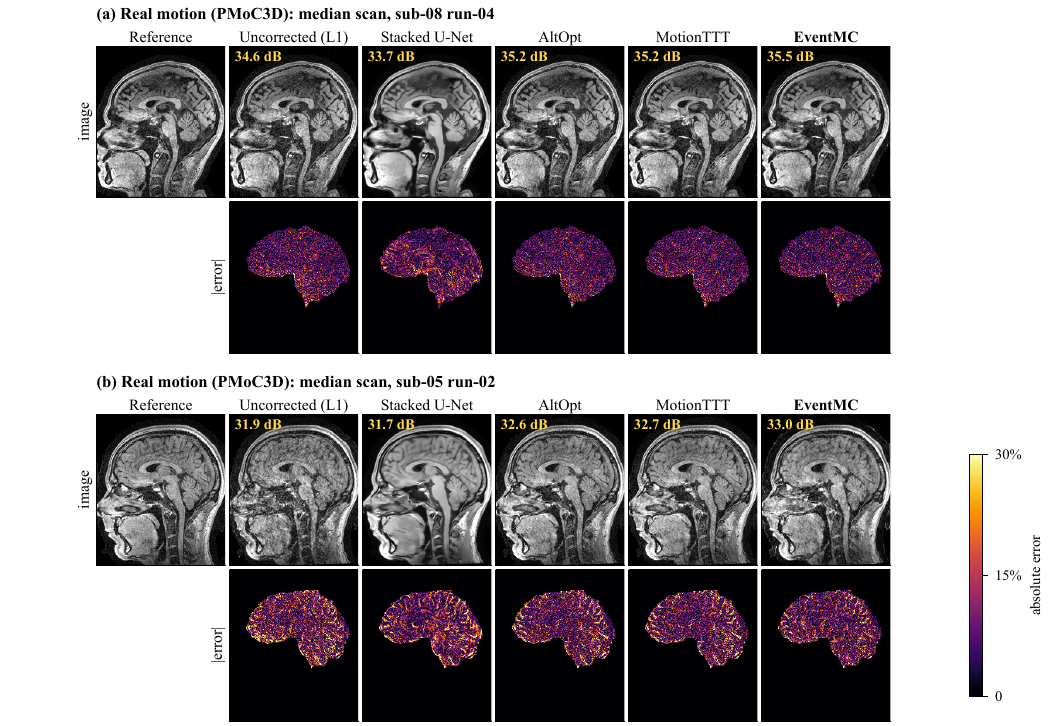}
\caption{Real motion (PMoC3D), central sagittal slices of the two median scans by per-scan PSNR difference between EventMC and MotionTTT. PSNR as in Table~\ref{tab:main}(b). Bottom rows: absolute error within the brain mask, relative to the normalised intensity range (Appendix~\ref{app:eval}).}
\label{app:fig-pmoc-median}
\end{figure*}

\begin{figure*}[p]
\centering
\includegraphics[width=\textwidth]{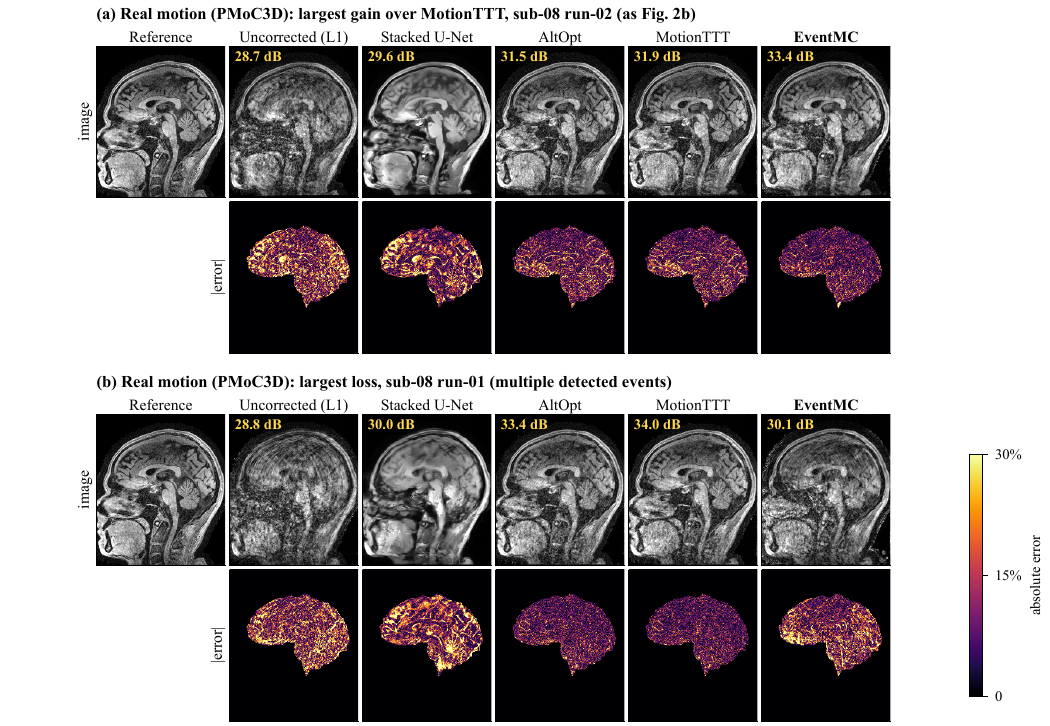}
\caption{Real motion (PMoC3D), central sagittal slices: (a) the scan with the largest gain over MotionTTT, also shown in Fig.~\ref{fig:cases}b, and (b) the scan with the largest loss, with multiple detected events. PSNR as in Table~\ref{tab:main}(b). Display as in Fig.~\ref{app:fig-pmoc-median}.}
\label{app:fig-pmoc}
\end{figure*}

Figs.~\ref{app:fig-accel}--\ref{app:fig-pmoc} show central sagittal slices with the same layout: the reference, each method with its volumetric PSNR, and the absolute error below. The cases are fixed by rule rather than chosen for appearance:
\begin{itemize}
\item \emph{Acceleration} (Fig.~\ref{app:fig-accel}): the case of Fig.~\ref{fig:cases}a at R2 and R4. At R2, EventMC has the smallest errors; at R4, MotionTTT, MotionTTT+TV and EventMC reach similar quality.
\item \emph{Sustained mid-scan motion} (Fig.~\ref{app:fig-m}): at R1 and at R2, the case whose PSNR gain of EventMC over MotionTTT+TV is the median of the nine subjects.
\item \emph{Real motion, typical} (Fig.~\ref{app:fig-pmoc-median}): the two PMoC3D scans in the middle of the per-scan PSNR differences from MotionTTT (13th and 14th of 26; $+0.27$ and $+0.35$~dB).
\item \emph{Real motion, extremes} (Fig.~\ref{app:fig-pmoc}): the PMoC3D scans with the largest gain and the largest loss relative to MotionTTT (Appendix~\ref{app:real}).
\end{itemize}

 \end{document}